\documentclass[sigconf]{acmart}

\AtBeginDocument{%
  \providecommand\BibTeX{{%
    \normalfont B\kern-0.5em{\scshape i\kern-0.25em b}\kern-0.8em\TeX}}}
    
\usepackage[utf8]{inputenc} % allow utf-8 input
\usepackage[T1]{fontenc}    % use 8-bit T1 fonts
\usepackage{hyperref}       % hyperlinks
\usepackage{url}            % simple URL typesetting
\usepackage{booktabs}       % professional-quality tables
\usepackage{amsfonts}       % blackboard math symbols
\usepackage{nicefrac}       % compact symbols for 1/2, etc.
\usepackage{microtype}      % microtypography
\usepackage{lipsum}		% Can be removed after putting your text content
\usepackage{graphicx}
\usepackage{natbib}
\usepackage{doi}
\usepackage{booktabs} % For formal tables
\usepackage{subcaption}
\usepackage{graphicx}
\usepackage{diagbox}
\usepackage{tikz}
\usepackage{tabularx}
\usepackage{fp}
\usepackage{colortbl}
\usepackage{collcell}

\usepackage{xcolor} % http://www.ctan.org/tex-archive/macros/latex/contrib/xcolor
\usepackage{tikz}

\newcommand\circlered[3][]{%
  \tikz[remember picture,baseline=(#2.base)]
    \node[minimum size=0pt,inner sep=0pt,#1](#2){#3};%
}
\usetikzlibrary{shapes.geometric, arrows, positioning, automata,positioning,fit,backgrounds}
\tikzstyle{process} = [rectangle, minimum width=2cm, minimum height=1cm, text centered, draw=black, fill=white!30]
\tikzstyle{sum} = \tikzstyle{sum} = [draw, circle, minimum size=.5cm]
\tikzstyle{arrow} = [thick,->,>=stealth]

\usepackage{hyperref}       % hyperlinks

\usepackage{amsmath}

\usepackage{bm}
\usepackage{units}
\usepackage{float}
\usepackage{dblfloatfix}
\usepackage[ruled,vlined]{algorithm2e}
\definecolor{lime}{HTML}{A6CE39}

\copyrightyear{2024}
\acmYear{2024}
\setcopyright{rightsretained}
\acmConference[GECCO '24]{Genetic and Evolutionary Computation Conference}{July 14--18, 2024}{Melbourne,  AUS}
\acmBooktitle{Genetic and Evolutionary Computation Conference (GECCO '24), July 14--18, 2024, Melbourne, AUS}
\acmDOI{----------}
\acmISBN{----------}
\begin{document}

%%
%% The "title" command has an optional parameter,
%% allowing the author to define a "short title" to be used in page headers.
% \title{Collective motion emerging from evolving swarm controllers in different environments using gradient following task}
\title{Emergence of specialized Collective Behaviors in Evolving Heterogeneous Swarms}

%%
%% The "author" command and its associated commands are used to define
%% the authors and their affiliations.
%% Of note is the shared affiliation of the first two authors, and the
%% "authornote" and "authornotemark" commands
%% used to denote shared contribution to the research.
\author{Fuda van Diggelen}
% \authornotemark[\orcidFuda]
% \email{fuda.van.diggelen@vu.nl}
\affiliation{%
  \institution{Vrije Universiteit Amsterdam}
  \streetaddress{De Boelaan 1111}
%   \city{Amsterdam} 
  \state{Noord-Holland} 
  \postcode{1081 HV}
  \country{The Netherlands}
}
\email{fuda.van.diggelen@vu.nl}

\author{Matteo De Carlo}
% \authornotemark[\orcidMatteo]
%\orcid{xxxx-xxxx-xxxx}
\affiliation{%
  \institution{Vrije Universiteit Amsterdam}
  \streetaddress{De Boelaan 1111}
%   \city{Amsterdam} 
  \state{Noord-Holland} 
  \postcode{1081 HV}
  \country{The Netherlands}
}
\email{m.decarlo@vu.nl} 

\author{Nicolas Cambier}
% \authornotemark[\orcidNicolas]
%\orcid{xxxx-xxxx-xxxx}
\affiliation{%
  \institution{Vrije Universiteit Amsterdam}
  \streetaddress{De Boelaan 1111}
%   \city{Amsterdam} 
  \state{Noord-Holland} 
  \postcode{1081 HV}
  \country{The Netherlands}
}
\email{n.p.a.cambier@vu.nl}

\author{Eliseo Ferrante}
\affiliation{%
  \institution{Vrije Universiteit Amsterdam}
  \streetaddress{De Boelaan 1111}
%   \city{Amsterdam}
  \state{Noord-Holland} 
  \country{The Netherlands}
}

\email{e.ferrante@vu.nl}

\author{A.E. Eiben}
\affiliation{%
  \institution{Vrije Universiteit Amsterdam}
  \streetaddress{De Boelaan 1111}
%   \city{Amsterdam} 
  \state{Noord-Holland} 
  \postcode{1081 HV}
  \country{The Netherlands}
}
\email{a.e.eiben@vu.nl}

%%
%% By default, the full list of authors will be used in the page
%% headers. Often, this list is too long, and will overlap
%% other information printed in the page headers. This command allows
%% the author to define a more concise list
%% of authors' names for this purpose.

%%
%% The abstract is a short summary of the work to be presented in the
%% article.
\begin{abstract}
%Genotypical variation inside a swarm of animals can lead to specific sub-group behavior when certain task specialization is advantageous. 
Natural groups of animals, such as swarms of social insects, exhibit astonishing degrees of task specialization, useful to address complex tasks and to survive.
This is supported by phenotypic plasticity: individuals sharing the same genotype that is expressed differently for different classes of individuals, each specializing in one task.
% In this work we study the evolution of task specialization through phenotypic plasticity in a swarm of simulated robots.
In this work, we evolve a swarm of simulated robots with phenotypic plasticity to study the emergence of specialized collective behavior during an emergent perception task.
Phenotypic plasticity is realized in the form of heterogeneity of behavior by dividing the genotype into two components, with one different neural network controller associated to each component. 
The whole genotype, expressing the behavior of the whole group through the two components, is subject to evolution with a single fitness function.
%We consider the system without and with a regulatory mechanism able to switch between these two components.
We analyse the obtained behaviors and use the insights provided by these results to design an online regulatory mechanism.
Our experiments show three main findings: 1) The sub-groups evolve distinct emergent behaviors. 
2) The effectiveness of the whole swarm depends on the interaction between the two sub-groups, leading to a more robust performance than with singular sub-group behavior. 3) The online regulatory mechanism enhances overall performance and scalability.
\end{abstract}

%%
%% The code below is generated by the tool at http://dl.acm.org/ccs.cfm.
%% Please copy and paste the code instead of the example below.
%%
\begin{CCSXML}
<ccs2012>
<concept>
<concept_id>10010520.10010553.10010554.10010556.10011814</concept_id>
<concept_desc>Computer systems organization~Evolutionary robotics</concept_desc>
<concept_significance>500</concept_significance>
</concept>
</ccs2012>
\end{CCSXML}

\ccsdesc[500]{Computer systems organization~Evolutionary robotics}

%%
%% Keywords. The author(s) should pick words that accurately describe
%% the work being presented. Separate the keywords with commas.
\keywords{Swarm robotics, Evolutionary robotics, Heterogeneous swarm}

%% A "teaser" image appears between the author and affiliation
%% information and the body of the document, and typically spans the
%% page.
% \begin{teaserfigure}
%   \includegraphics[width=\textwidth]{sampleteaser}
%   \caption{Seattle Mariners at Spring Training, 2010.}
%   \Description{Enjoying the baseball game from the third-base
%   seats. Ichiro Suzuki preparing to bat.}
%   \label{fig:teaser}
% \end{teaserfigure}

%%
%% This command processes the author and affiliation and title
%% information and builds the first part of the formatted document.
\maketitle

\section{Introduction}
Collective motion is widely documented in groups of animals in nature and has shown to enhance the group with capabilities that are not apparent in an individual group member. Those emergent capabilities include increased environmental awareness \cite{couzin2005-animalgroups,kearns2010-bacterialswarming}, protection against predators \cite{olson2013-predatorconfusion}, and gradient sensing \cite{puckett2018collective}. Swarm robotics aims at implementing such collective behaviors in robots in order to leverage those advantageous emergent properties for engineering purposes. In particular, collective motion can enable robotic swarms to achieve sensing beyond the capabilities of the sensors fitted to individual agents \cite{karaguzel2020collective}, i.e. emergent perception. 

The main challenge in designing collective motion behaviors, and swarm robotics systems in general, comes from the fact that designers can only implement robot controllers on individual robots, but that the desired behavior is defined at the group level \cite{hasselmann2021empirical}.
Therefore, successful group behaviors depend on emergent properties, which are difficult to predict. 
An approach to this problem is to define a success metric at the group level and to use it as a reward to automatically optimize robot controllers \cite{trianni2014evolutionary,francesca2014automode,hasselmann2021empirical, van2022environment}. 
However, the solutions that emerge through this process tend to overfit their training environment and, consequently, lack flexibility. 
Therefore, automated design requires a framework that maintains good performance under a variety of environmental conditions. For example, modularization of individual-level control has been proposed as a solution to increase flexibility \cite{francesca2014automode}.

We believe that, by leveraging on heterogeneity, we can achieve a modular framework at the swarm level, as opposed to the prevailing `homogeneous designs' typically found in the swarm robotics literature \cite{trianni2014evolutionary,francesca2014automode,hasselmann2021empirical}. 
Heterogeneous collective behaviors are widely observed in nature. 
For example, social insects divide their tasks into sub-tasks (the so-called division of labor), assigned to specific group members, in order to improve their efficiency \cite{ratnieks1999task}. 
Indeed, distinct sub-group behaviors can emerge when task specialization is beneficial for the group as a whole \cite{ioannou2017swarm}. 
Maintaining a `group identity' whilst splitting in sub-groups can be achieved through genetic homogeneity with phenotypic heterogeneity.
For example, within an insect colony, members share similar-to-identical genotypes \cite{ravary2007individual}; therefore, the various behaviors used by insects are all encoded in the same genotype and are activated by external cues, e.g. queen pheromones \cite{holman2018queen}. 
This phenomenon, in which the same genotype expresses a different phenotype by gene regulatory mechanisms, is known as \textit{phenotypic plasticity}. 
Although it is more commonly useful for task partitioning \cite{ratnieks1999task, ferrante2015evolution}, it has also been found in species exhibiting collective motion \cite{ariel2015locust}.

% In this paper, we propose a novel framework that promotes modularity at the swarm level through heterogeneity, to improve the flexibility of automated designs in self-organized collective behaviors.
In this paper, we propose to improve the flexibility of automated designs by promoting modularity at the swarm level through heterogeneity in self-organized collective behaviors (as apposed to individual-level).
Our framework is based on phenotypic plasticity, where we evolve separate controllers for sub-groups inside the swarm and, through a regulatory mechanism, adjust the phenotypic ratio of each group. 
% Previous works on this matter focused on adjusting this ratio \cite{brutschy2014self,valentini2022global} or on the specific problem of task partitioning \cite{tuci2015design,ferrante2015evolution}, i.e. allocating robots to sequential or parallel tasks, and enabling the appropriate behavior in the ensuing sub-groups. 
More specifically, we consider polyphenism, a specific case of phenotypic plasticity, whereby phenotypic plasticity is expressed at birth and remains constant throughout life \cite{ariel2015locust}. Our work is novel as it evolves an heterogeneous swarm on a swarm level without a priori knowledge of how heterogeneity should be leveraged in the swarm. This makes our method task-agnostic and highly adaptable for other tasks. 

The paper is organized as follows. In \autoref{sec:rel_work}, we present the state of the art on automated design for heterogeneous swarms, and outline our unique approach, %that enables the automatic design of specialized behaviors in swarm members to achieve a singular task, namely
which we on an emergent perception task. 
Our implementation is detailed in \autoref{sec:meth}, considering robots that have limited sensing capabilities (which prevents them from achieving the group task individually) and lack awareness of the specific roles or specializations within the swarm.
In \autoref{sec:res}, we present our optimization results and re-test our best controller with different sub-group ratios. This analysis provides insights that we use to design an online regulatory mechanism dependent on local conditions, where robots automatically switch between the controllers using a probabilistic finite-state machine. %These results and the phenotypical plasticity rules are presented in the results, .
We discuss, in \autoref{sec:disc}, whether specialization and cooperation can arise between the different sub-groups, without the need for explicit rewards to encourage these behaviors
Finally, we present our final remarks and future work in \autoref{sec:conc}.

%In order to test the complementary of the emerged specializations, we study the evolved controllers without a regulatory mechanism, that is with different fixed ratios of robots equipped with each controller, as well as with a regulatory mechanism based on probabilistic finite state machines that enables robots to automatically switch between the controllers. %which highlights our framework's potential for flexibility.

% In the end, we are interested in the following:

% \begin{itemize}
%     \item Will distinct sub-group behavior emerge?
%     % \item Is specialized behavior beneficial with respect to optimizing a homogeneous swarm?
%     \item Is there a vital interaction between sub-groups that makes their combination more viable?
% \end{itemize}

\section{Related work}\label{sec:rel_work}

Behavioral heterogeneity induces several challenges, the first of which is task allocation, i.e. dynamically adjusting the number of agents assigned to each available task \cite{bayindir2016review}. The problem of achieving this goal in a decentralized manner has been widely addressed by swarm robotics. Approaches to this problem usually focus on the mechanism of task switching and, therefore, use relatively simplistic, manually implemented behaviors. 
With threshold-based responses, robots initially show some preference for a given task, while simultaneously recognizing deficiencies in the accomplishment of other tasks (e.g. objects accumulating). Beyond some threshold, the robot switches to the corresponding task \cite{krieger2000call}. 
In some cases, this switch is probabilistic in order to avoid large-scale population switches, which might leave another task unaddressed \cite{brutschy2014self}. Mathematical modeling of task allocation has also been proposed \cite{valentini2022global}, which allows custom task allocation parameters according to their needs. 

In addition to task allocation, task specialization focuses on the emergence of several complementary functions within the swarm. It often refers to physically heterogeneous swarms, i.e. multi-robot systems composed of multiple robot platforms \cite{rizk2019cooperative}. It should be noted that this impairs a vital advantage of swarms, namely robustness, as in such cases the individual members are not interchangeable with one another. A more robust design is when agents are physically identical but differ in behavioral function \cite{bettini2023heterogeneous}.
Functional heterogeneity through behavioral specialization does not suffer from this problem \cite{hussein2022autonomous}. Here, it is possible for robots with the same body to switch behavior while employed, also called task partition. In such a context, maintaining collective behavior at the group level can be tricky, as online regulatory mechanisms tend to switch individual behaviors only \cite{feola2023aggregation}.

Tuci et al. investigated the evolution of task partition (i.e. both task allocation and specialization), with a \textit{clonal} and \textit{aclonal} evolutionary process, in a physically homogeneous swarm of five e-pucks. Here, \textit{clonal} refers to a single genotype being shared by all the agent, whereas, with \textit{aclonal}, their genotypes are all different from each other. Unsurprisingly, they found that robots performed better in their \textit{aclonal} approach \cite{tuci2013evolution, tuci2014evolutionary}, especially if their controller was optimized with a multi-objective fitness \cite{tuci2015design}, as they could address the required sub-tasks in parallel. Notably, this approach results in an efficient optimization process, as each agent in the swarm samples a different genome. Moreover, plasticity was observed, in the sense that individual robots were able to switch tasks according to environmental requirements (including their peers' behaviors).

Closely related, \cite{ferrante2015evolution} also addressed task partition, with a task that evokes the environmental context of leafcutter ants who divide their foraging task into two sub-tasks: cutting and dropping leaf fragments into a storage area, on the one hand, and collecting and bringing the fragments back to the nest, on the other. Their experimental environment was composed of a slope separating a nest (below) and a source (above) area so that the robots could individually retrieve the food objects back to the nest, or deposit them on the slope and relying on other robots to fetch them. Task allocation was evolved on a probabilistic finite-state machine, composed of simplistic pre-programmed behaviors, without specifying a preference for collective, rather than individual behavior. Their success demonstrates that even homogeneous controllers (i.e. a single phenotype) can handle task partition through individual experience, stigmergy, and stochastic switching alone, given enough knowledge on the task to design viable behaviors. 

The presented work on heterogeneous swarm optimization requires in-depth knowledge of the specific learning task. Whether it be the design of sub-tasks/goals, pre-formulized (modular) behaviors, or finite states; a priori knowledge is required for the design of these controllers. 
We aim to make collective specialized behavior emerge, without any explicit reward on the specific sub-tasks. Differently from the aforementioned works, our approach require minimal insight about the solution as we do not pre-define various subtasks/behaviors, such as bucket brigading, finite-state machines or task-allocations. Instead, we define a reward on the overall group-level performance of the whole task and let specialization evolve as an optimal solution. This simplifies the design of a good performance metric, as it alleviates the necessity to define a priori what `good' specialized behavior constitutes. 
Furthermore, our method is more flexible than a modular design with pre-defined specialized behaviors that presuppose sub-group interactions affecting the overall swarm performance. Additionally, we automatically obtain task allocation through a heuristic approach that designs our online regulatory mechanism. This straightforward method minimizes the time and effort required, while demonstrably enhancing overall task performance. Altogether, this work shows that optimal task partition can be designed automatically using evolutionary computing, without any specific sub-task knowledge. To the best of our knowledge, no previous work exists that addresses task specialization and allocation in such a context.

\section{Methodology}\label{sec:meth}
% Our framework is based on \cite{van2022environment}, which showed that collective behavior can emerge from optimization on task performance, rather than optimization on the group behavior itself. For this purpose an evolutionary algorithm and Reservoir Neural Network (RNN, \cite{lukovsevivcius2009reservoir}) were used. RNN are neural networks with fixed weights on all edges, except those connecting to the output layer. This results in a smaller number of variables, which makes evolution more efficient.
Optimising a heterogeneous swarm controller without specific knowledge on any sub-task requires a flexible approach that can be applied on any type of task. We test our method in an emergent perception task for gradient sensing where robots have to find the brightest spot in the center (inspired by fish behavior described in \cite{puckett2018collective}). For this, we utilize black box optimization in the form of an evolutionary algorithm on Reservoir Neural Networks (RNN), a method that has been employed on homogeneous swarms in a similar capacity \cite{van2022environment}. 
Full code base can be found in the following git repository after acceptance: 
% \href{https://github.com/fudavd/EC_swarm/tree/AAMAS_2024}{https://github.com/fudavd/EC\_swarm/tree/AAMAS\_2024} 
\href{ANONYMOUS}{<ANONYMOUS placeholder>}

\subsection{Robot design}
Each robot in the swarm has an identical differential drive hardware design based on the Thymio II without any communication capabilities (Bluetooth, radio, WiFi). Our robot consists of a cart with two actuated wheels (max speed is $\pm \unit[14]{cm/s}$) in the back and a single omni-directional wheel in the front. We equip our robots with range and bearing sensing in 4 directions (specifics are detailed in \autoref{sec:Control}) and a local value sensor to measure local light intensity. These sensors are sampled at 10Hz to obtain control inputs only based on current information, i.e. no memory of previous state. 

It is important to be explicit on the capabilities of our robot design: 1) Robots do not communicate information to each other (e.g. local values at their position, future motor inputs, or any form of message passing); 2) The controller is memoryless, only current local sensor readings are known; 3) there is no notion of specialization inside the controller, meaning robots are `unaware' of specialization inside the swarm. All in all, this grounds the idea of `limited sensing', as a single robot is incapable of estimating the gradient of the light.

\subsection{Controller design}\label{sec:Control}
For general applicability, we require our controller to be as flexible as possible while capable of learning quickly. For this we opt to use neural networks (which are expressive function approximators) with random functionalities in the form of a reservoir to speed up learning (i.e. RNN, \cite{lukovsevivcius2009reservoir}). This reservoir is created by freezing the network weights up to the last layer after random initialization, resulting in a fixed set of functions from which we learn an optimal combination in the final network layer. %Our RNN has the following inputs: relative distance and heading information from 4 directional sensors, plus a local value sensor reading (totaling 9 different inputs). 

To allow specialization, we divide our swarm into two sub-groups, with each sub-group containing a different RNN controller (all sub-group members have the same RNN). The two RNNs are randomly initialized with different reservoirs that we save at the start. We describe the swarm genotype as a single vector of weights from which the first half refers to the last layer of the first RNN, and the second half to the last layer of the second RNN. The phenotypic plasticity of our single genome is expressed through our sub-group division.

\begin{figure}[hpt!]
  \includegraphics[width=0.9\linewidth]{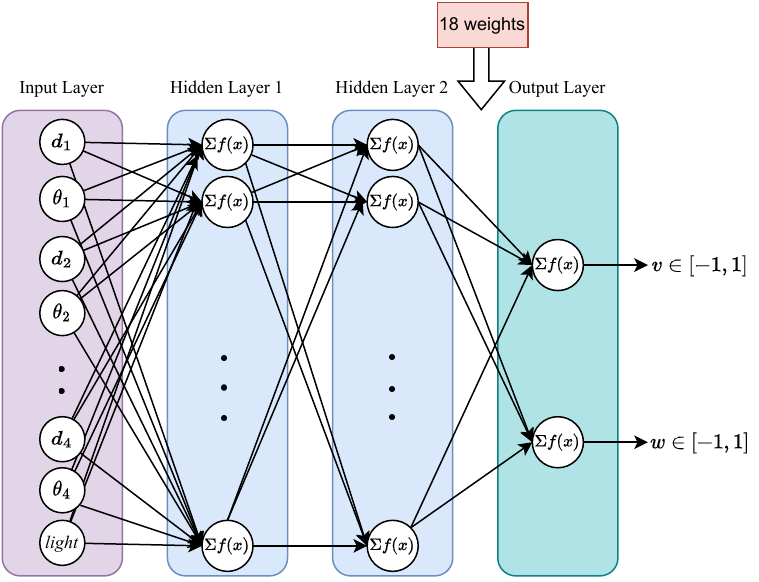}
  \centering
  \caption{\small Reservoir Neuron Network controller design.}
  \label{fig:NN}
\end{figure}

The phenotype of the controller is illustrated in \autoref{fig:NN}. The RNN has an input layer of 9 neurons that are rescaled to [-1, 1], namely 4 directional sensors (each providing two values: a distance and a heading) and 1 local value sensor. The 4 directional sensors cover a combined 360-degree view of the robot's surroundings (front-, back-, left-, and right quadrants of 90 degrees each). Within each quadrant ($i$) the sensor obtains the distance ($d_i$) and relative heading ($\theta_i$) of the nearest neighbor up to a maximum range of 2 meters (outside of this range the sensor defaults $d_i$=2.01 and $\theta_i$=0). The RNN outputs target speed ($v\in[-1,1]$) and angular velocity ($w\in[-1,1]$), which are transformed into direct velocity commands for the two wheels.

The RNN architecture is a fully connected neural network with an input layer with 9 neurons ($\mathbf{s}_{in} \in [-1,1]^9$), 2 hidden ReLU layers of the same size (${h}_{1}, {h}_{2} \in \mathbb{R}^9$), and a final output layer with two tanh$^{-1}$ neurons, $RNN \in [-1,1]^2$. All hidden reservoir weights are initialized randomly with a uniform distribution ($U[-1,1]$). We set all biases to 0 and only optimize the weights of the output layer during evolution (18 weights per RNN). The final RNN controller can be formalized as follows:

\begin{align*}
    RNN = \mathrm{tanh}^{-1}\left(\mathbf{W}_{out}\mathrm{ReLU}\left(\mathbf{W}_{h2}\mathrm{ReLU}\left(\mathbf{W}_{h1}\mathbf{s}_{in}\right)\right)\right) \\
    \text{$\textrm{with,} \quad  \mathbf{W}_{h1, h2} \in \mathbb{R}^{9\times9} \quad \textrm{and} \quad \mathbf{W}_{out} \in \mathbb{R}^{2\times9}$ }
\end{align*}

\subsection{Emergent perception task}
We aim to enhance the sensing capability of our robots so that, when operating collectively as a swarm, they can perceive the gradient. Our set-up consist of 20 robots that are rewarded for navigating to the brightest spot in the center of a 30x30m arena. Since an individual robot lacks the capacity to sense the direction of the gradient field, we anticipate that a collective behavior will emerge. This `task' is not too dissimilar to a behavior found in schools of fish that tend to aggregate in the shadow as it decreases their visibility from predators \cite{puckett2018collective}. 

\begin{figure}[ht]
\vspace{-0.5em}
  \centering
\begin{minipage}[c]{0.95\linewidth}
\centering
        \begin{minipage}[t]{.4\linewidth}%## ROW 10x10
            \centering
            \subfloat[(a) Scalar field map]{\includegraphics[width=\textwidth]{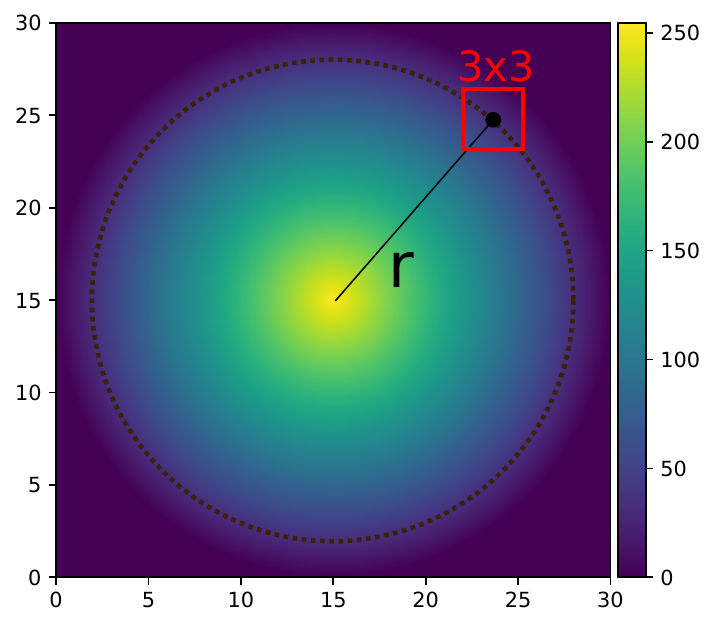}}
        \end{minipage}
\hspace{2em}
        \begin{minipage}[t]{.4\linewidth}
            \centering
            \subfloat[(b) Isaac gym]{\includegraphics[width=\textwidth]{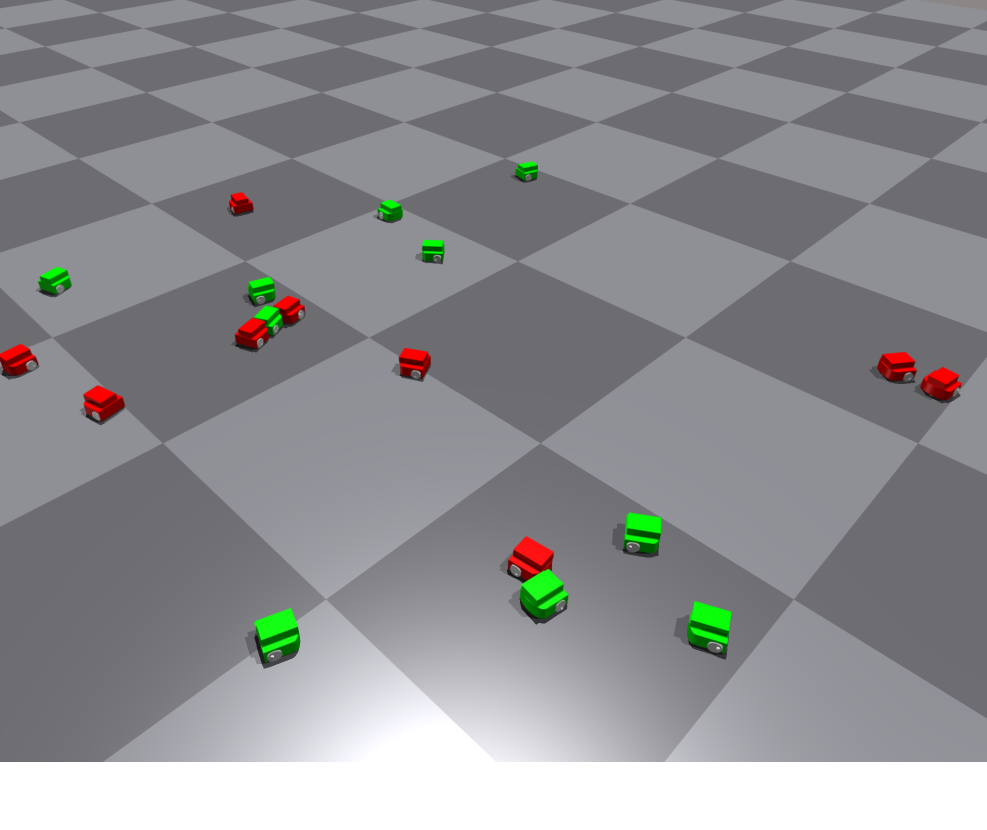}}
    \end{minipage}
\end{minipage}\vspace{-1em}
\caption{\small The task environment. (a) The scalar map indicating a random instance of our swarm task. (b) Our swarm with different sub-groups (colored red and green) in simulation.}
\label{fig:swarm_env}
\vspace{-1em}
\end{figure}

We use Isaac Gym from \cite{makoviychuk2021isaac} to simulate our swarm(s) ($\mathrm{dt=0.05s}$). The task environment is a scalar field map with its maximum value (255) in the center (see \autoref{fig:swarm_env}a). We randomly placed the swarm in a circle at a fixed distance (r=12m) away from the center. At this position, we randomly place each swarm member within a 3x3m bounding box (shown in red). The swarm is split into two sub-groups (red and green) of 10 members each, as shown in \autoref{fig:swarm_env}b. 

\begin{figure*}[h] 
  \centering
  \includegraphics[width=400pt, trim={0 35px 0 105px},clip]{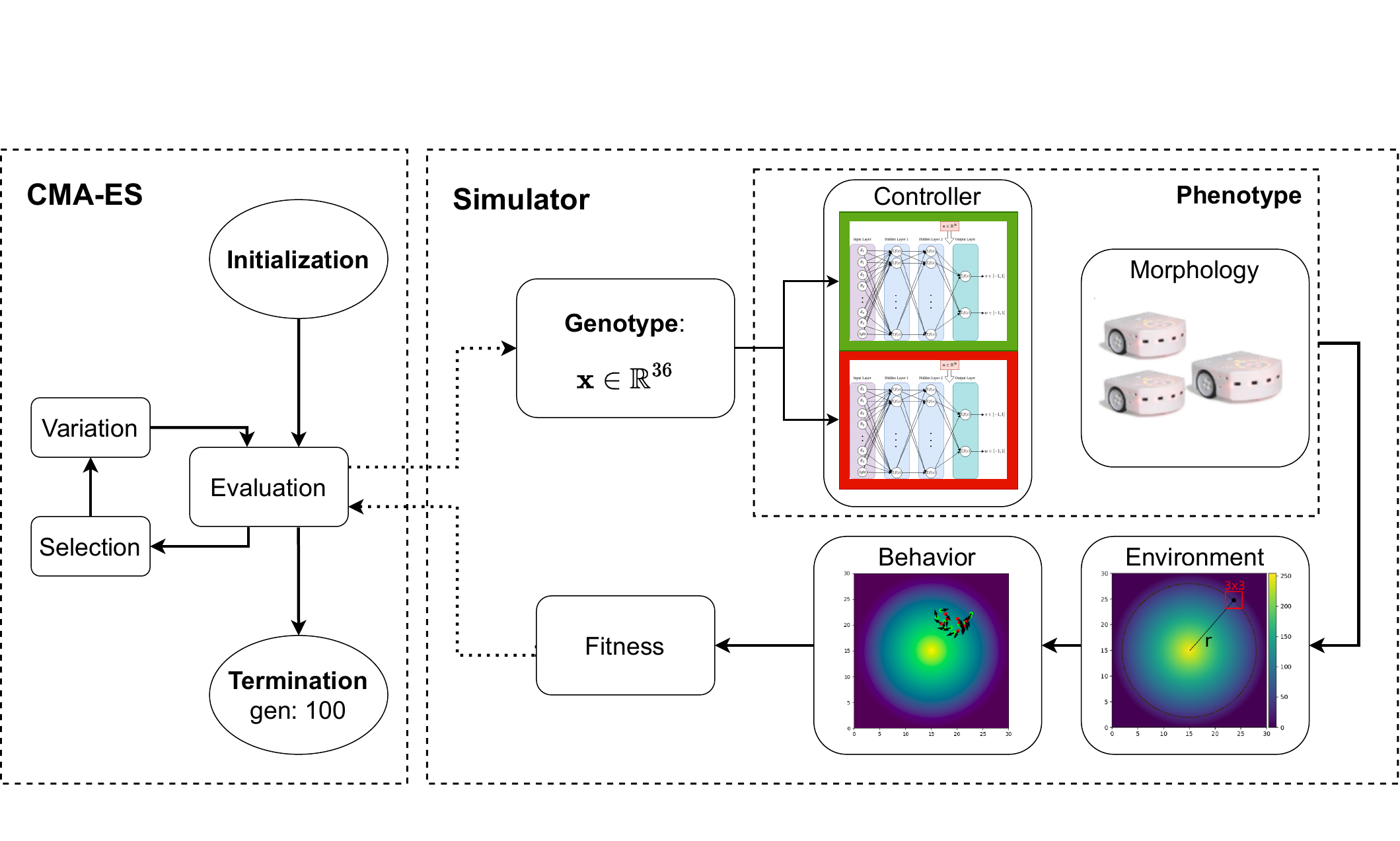}
  \vspace{-0.5em}
  \caption{\small Experimental setup for optimizing swarm controllers, where an evolutionary algorithm evaluate different genotypes in our swarm simulator (big dashed box). Please note that our genotype encodes two different controllers colored green and white boxes.}
  \label{fig:whole_process}\vspace{-1em}
\end{figure*}

\subsection{Evolving swarm experiment}
We optimize our swarm using the Covariance Matrix Adaptation Evolution Strategy (CMA-ES, \cite{hansen2001cmaes}). We run 10 repetitions of our experiments to obtain the overall best controller.

Let us draw a clear distinction between the components within \textit{CMA-ES}, where individuals in a population are being optimized, and our \textit{swarm} which refers to an instance of an individual, consisting of robots/members that are assigned to different a sub-group (see \autoref{fig:whole_process}).
Within our evolutionary algorithm, we thus have individuals that we want to evaluate.
All individuals within this evolving population have two RNNs with the same two reservoirs.
Differences between individuals are defined by their genotype (a vector of 36 weights, $\textbf{x} = \left[\mathbf{W}_{out_{1:}}, \mathbf{W}_{out_{2:}}\right]$ with, $\textbf{x} \in \mathbb{R}^{36}$) that encode the last layer weights of the two RNNs. 
We evaluate our individual by assigning the RNNs to two sub-groups in a single swarm of robots, where each sub-group member (i.e. a robot belonging to a specific sub-group) has the same RNN as the other constituents. After a trial we calculate a fitness value based on task performance of the swarm and assign it to the corresponding individual. 
% We evaluate every individual in the population of swarms until we can do selection and mutation. 

CMA-ES is a sampling-based evolutionary strategy that aims to find a distribution in the search space to sample high-performing individuals with high probability. % Pseudo-code is shown below \autoref{alg:cmaes}, actual implementation is in python using \cite{nikolaus_hansen_2023_7573532}. 
Here, CMA-ES samples new candidates $\textbf{x}$ according to a multivariate normal distribution. 
The covariance matrix of the sampling distribution is updated at each generation ($\textbf{C}_{gen}$) to increase the likelihood of sampling an individual with higher fitness. 
% \begin{algorithm}[h]
% \SetAlgoNlRelativeSize{0}
% \caption{\small CMA-ES Pseudocode\label{alg:cmaes}}
% \KwData{Initial mean $\mathbf{m}_0 = 0$, initial covariance $\sigma_0$, population size $\lambda = 30$ solutions $\{\mathbf{x}_1, \mathbf{x}_2, \ldots, \mathbf{x}_\lambda\}$ with genotype $\mathbf{x}_i \in \mathbb{R}^{36}$}
% \KwResult{Optimal solution $\mathbf{x}^*$}
% % $gen \leftarrow 0$

% $N_{gen}\leftarrow 100$

% $N_{repeats}\leftarrow 3$

% \For{$gen$ in $N_{gen}$}{
%   \For{$\mathbf{x}_i$ in  $\lambda$}{    
%     \For{$rep$ in $N_{repeats}$}{
%         $f_{rep} \leftarrow \textrm{SIMULATE}(\mathbf{x}_i)$ \hfill {\textit{simulate for }10min}
%     }

%     $F_i\leftarrow \textrm{MEDIAN}(f_1, \dots, f_{N_{repeats}})$ \hfill \textit{Final fitness}
%   }
%   Sort solutions by fitness
  
%   Select the top-performing solutions to form a weighted mean
  
%   $\mathbf{m}_{t+1} \leftarrow \sum_{i=1}^{w} w_i \mathbf{x}_i$, where $w_i$ are weights
  
%   Update covariance matrix $\mathbf{C}_{gen+1}$
  
%   Generate a new population covariance matrix $\mathbf{C}_{\text{new}}$ based on the selected solutions
  
%   $\mathbf{C}_{gen+1} \leftarrow \mathbf{C}_{\text{new}}$
  
%   Sample $\lambda$ solutions $\{\mathbf{x}_1, \mathbf{x}_2, \ldots, \mathbf{x}_\lambda\}$ using $\mathbf{C}_{gen+1}$
% }
% $\mathbf{x}^* \leftarrow$ Best solution found
% \Return{$\mathbf{x}^*$}
% \end{algorithm}
We set our population size to 30 individuals (i.e., 30 different swarms) and evolve for 100 generation. Every individual is tested three times, with the median running to represent the final fitness. This reduces the sensitivity to lucky runs that are nonrepeatable and therefore nonviable. 

\begin{table}[htp!]
\footnotesize
\caption{\small Evolving swarm experiment parameters}
\centering
% \vspace{-0.5em}
\vspace{-1.5em}
\begin{tabular}[b]{{p{0.19\linewidth} p{0.12\linewidth} p{0.53\linewidth}}}
                 & Value           & Description\\
\toprule
Runs      &  10             & Number of Runs of our experiment \\ 
% \bottomrule 
% \toprule
\midrule
\multicolumn{3}{l}{\textbf{Learning task}: \textit{collective gradient sensing}} \\
\midrule
Swarm size   & 20    & Number of robots in a swarm \\
Ratio        & 1:1   & sub-group division          \\ 
r        & 12   & Spawn distance from center (meters)          \\ 
Arena type  & center     & Environment: center	    \\
Eval. time   & 10    & Test duration in minutes    \\ 
% \bottomrule 
% \toprule
\midrule
\multicolumn{3}{l}{\textbf{Optimizer}: \textit{CMA-ES}}     \\
\midrule
genotype      & $U[-5,5]$ & Initial sampling $\mathbf{x}_{init} \in \mathbb{R}^{36}$   \\
$\lambda$   & 30        & Population size          \\
$N_{gen}$ & 100       & Termination condition            \\ 
$\sigma_0$  & 1.0       & Initial covariance value              \\ 
$N_{repeats}$     & 3         & Number of repetitions per individual \\
\bottomrule 
\end{tabular}
\vspace{-1.5em}
\label{tab:parameters}
\end{table}

\subsubsection{Fitness function}
We define the fitness of a swarm as generally as possible, solely based on the ability to follow the increasing gradient of the scalar field defined by the environment (shown in \autoref{fig:swarm_env}-a). This is done by aggregating the average light intensity values of all members over time (see \autoref{eq:r_i_sum}). 

\begin{equation}
{f} = \frac{\sum_{t=0}^{T}{l_t}}{G_{max}\cdot{T}}\quad\text{and}\quad{l}_t = \frac {\sum_{n=1}^{N}{G_n}}{N}
\label{eq:r_i_sum}
\end{equation}

Where $G_n$ is the scalar value of the grid-cell in which agent $n$ (of all agents, $N$) is located at a time $t$. Therefore, the fitness at a specific time ($l_t$) is calculated as the mean scalar light value of all swarm members. The trial fitness ($f$) is calculated by averaging all $l_t$ over total simulation time $T$. Finally, we normalize using the maximum scalar value $G_{max}$, always equal to $255$ for all experiments. A theoretical maximum fitness of 1 can only be achieved if all members of the swarm stack up in the center for the entire run. Fitness is only evaluated on swarm level with no distinction between sub-groups or any other task-specific principles that promotes specialization. Additionally, we would like to emphasize that our fitness function does not distinguish between robots that are sensing and following the increasing gradient collectively or as solitaries, or sub-groups that cooperate or not.

\subsection{Validation experiments}
After our evolutionary experiment, we obtain the overall best controller over 10 runs. We re-test this controller to test the emergent perception capability of the swarm to sense the gradient. Our validation is split in two: 1) We are particularly interested in the possible collective (sub-group) behavior(s) and the importance of sub-group interactions, which we elucidate by re-testing the best evolved controller and analysing the swarms behavior; 2) Using our two RNNs, we implement a straightforward online regulatory mechanism to mimic phenotypic plasticity induced by pheromones \cite{holman2018queen}. We investigate the viability of our (adaptive) heterogeneous swarm through scalability and robustness experiments.

\subsubsection{Collective behavior \& sub-group interactions}
First, we assess whether collective motion has emerged in a single re-test. We look at two different aspects of collective behavior over time: (1) \textit{performance} as mean scalar light value of the swarm (\autoref{eq:r_i_sum}, $l_t$); (2) \textit{alignment} in terms of order (\autoref{eq:align}, $\Phi$), which is defined as follows: 

\begin{equation}
\begin{aligned}
{\Phi} = \frac{\sum_{n=1}^{N}{\varphi_n}}{N}\quad 
\text{and}
\quad{\varphi_n}=\frac{\Big|\Big|\left(\sum_{p=1}^{P}{\angle{e^{j \theta_p}}}\right)+\angle{e^{j \theta_n}} \Big|\Big|}{P+1}
\end{aligned}
\label{eq:align}
\end{equation}

Here, $\varphi_n$ defines the order value calculated for agent $n$. Which is the average current heading direction of agent $n$ (noted as $\angle{e^{j \theta_n}}$) and all its perceived neighbors $P$ (noted as $\angle{e^{j \theta_p}}$). The total swarm order $\Phi$ is then defined as the average $\varphi_n$ over all agents in the swarm. The order measure gives a powerful insight into the alignment of agent's direction of motion. If all agents move towards the same direction, then the order measure approaches 1; and if they move in different directions, the order approaches 0.

Additionally, we investigate the benefits of sub-group interactions on two factors, namely performance and robustness. We retest our best controller in the same environment where we change the sub-group ratios (ratio $\in$ \{4:0, 3:1, 2:2, 1:3, 0:4\}). Different sub-group ratios can tell us if one sub-group is mainly responsible for the swarm performance or if sub-group interactions are important. These different ratios are tested by initializing the swarm at different distances from the center (r$_{ratio}$ $\in \left\{ 0,0.25,0.5,0.75,1\right\}$ as a ratio of the original training distance, 12m).

\subsubsection{Online regulatory mechanism}
Based on the results of the sub-group interactions experiment we can heuristically identify the best performing sub-group ratios at certain light intensities. Subsequently, we create a probabilistic finite state machine where the choice of sub-group controller within a robot is defined such that, on a holistic group level, the optimal sub-group ratios should emerge. This probability is only dependent on the local light value, and thus no communication is implemented to adapt to the best ratio. For example, if we heuristically find a 1:3 ratio to be the best sub-group division at the current local light intensity, the probability to sample an action from the second reservoir is 75\%. We update the probabilistic reservoir state every 5 seconds for stable behavior (this update frequency is found to be optimal by parameter sweep $\{1, 5, 10, 30, 60, 100\}$ seconds). 

\subsubsection{Scalability \& Robustness}
\begin{figure}[h]
\vspace{-0.5em}
  \centering
\begin{minipage}[c]{0.95\linewidth}
\centering
        \begin{minipage}[t]{.3\linewidth}%## ROW 10x10
            \centering
            \subfloat[(a) Bi-modal]{\includegraphics[trim={8em 2em 8em 1em},clip, width=\textwidth]{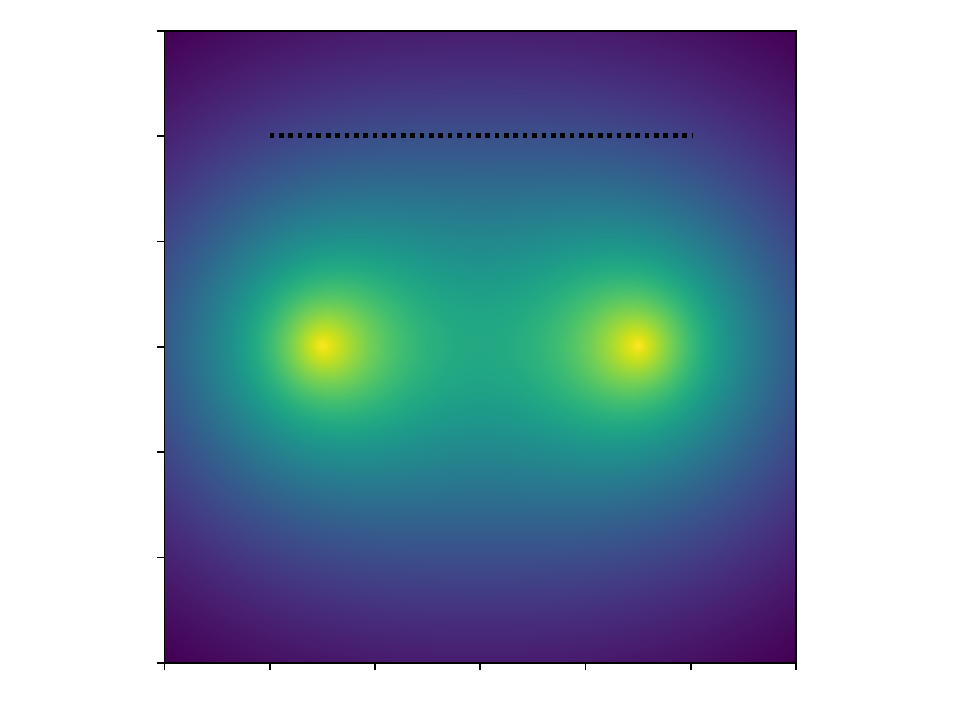}}
        \end{minipage}
% \hspace{2em}
        \begin{minipage}[t]{.3\linewidth}
            \centering
            \subfloat[(b) Linear]{\includegraphics[trim={8em 2em 8em 1em},clip,width=\textwidth]{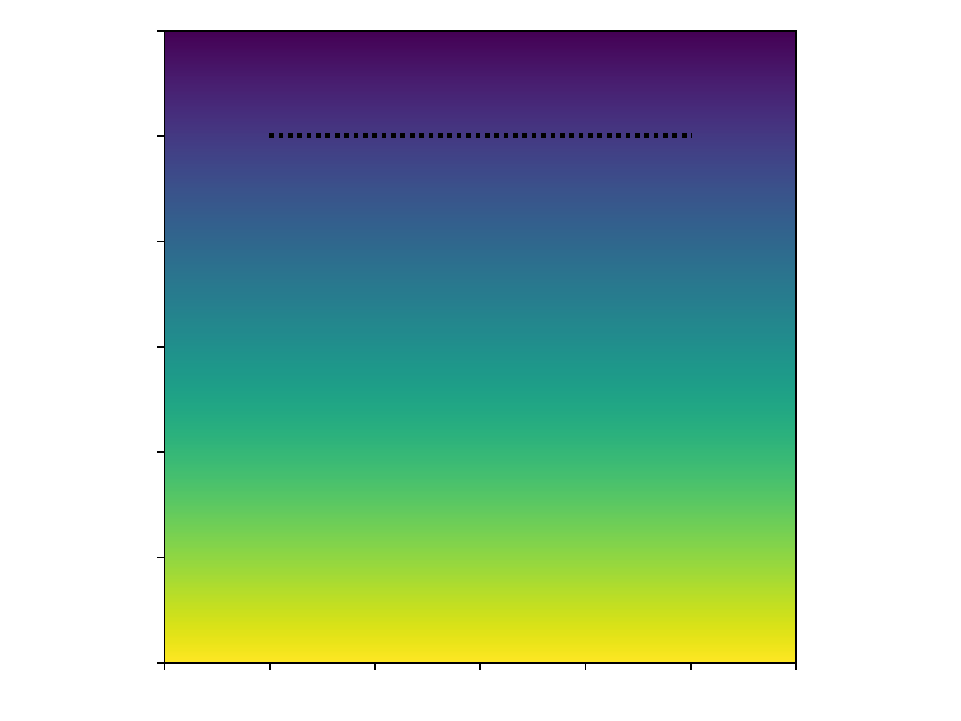}}
    \end{minipage}
% \hspace{2em}
        \begin{minipage}[t]{.3\linewidth}
            \centering
            \subfloat[(c) Banana]{\includegraphics[trim={8em 2em 8em 1em},clip, ,width=\textwidth]{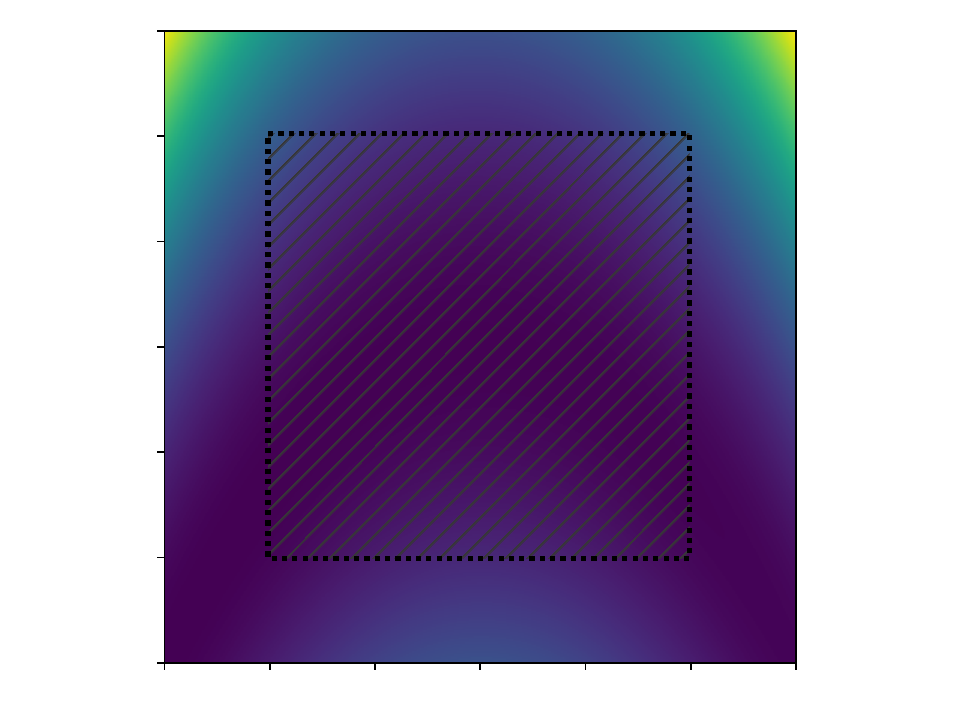}}
    \end{minipage}
\end{minipage}\vspace{-1em}
\caption{\small Validation environments. The black striped line indicates the random initialization location of the swarm (similar to \autoref{fig:swarm_env}). The striped box in (c) indicates the area of random initialization}
\label{fig:val_env}
\vspace{-1em}
\end{figure}

The swarm should be able to operate with a wide range of group sizes (i.e. Scalability) and across different types of environment (i.e. Robustness), using the same best controller. In our Scalability experiment, we initialize the swarm in the same environment but with the following swarm sizes [10, 20, 50] to see the impact on performance. In our robustness experiments, we initiate the swarm (of size 20) in different gradient maps ['Bi-modal', 'Linear', 'Banana'] as shown below. Each new arena poses different challenges: \textit{Bi-modal}, requires collective decision on where to go (\autoref{fig:val_env}a); \textit{Linear} poses a less salient gradient stretched over the full arena (\autoref{fig:val_env}b); \textit{Banana}, the banana function is a classic nonlinear minimization problem \cite{clerc1999swarm} with a curved shallow bottom. For our collective gradient ascent task, this function is interesting as it has both shallow gradient and local maxima (\autoref{fig:val_env}c). The experimental parameters we used in all the validation experiments are described in \autoref{tab:parameters_val}.

%In practice, it takes about 2.60 days to run these experiments on 5 computers. 

% \setlength{\tabcolsep}{10pt} % Default value: 6pt
\renewcommand{\arraystretch}{0.6} % Default value: 1
\begin{table}[h]
\footnotesize
\caption{\small Validation experiment parameters}
\centering
\vspace{-1em}
% \vspace{-2.5em}
\begin{tabular}[b]{{p{0.18\linewidth} p{0.20\linewidth} p{0.43\linewidth}}}
 & Value & Description\\
\toprule     
Repetitions      &  60    & Number of runs per experiment \\ 
Statistics       & \textit{t}-test & Statistical test \\ \bottomrule 
\toprule
\multicolumn{3}{l}{\textbf{Ratio}: \textit{sub-group division environment}} \\\midrule
Swarm size 			 & 20     & Number of robots in a swarm 	\\
Ratio    & 0:4/1:3/2:2 & sub-group division (+inverse) \\
r$_{ratio}$    & \tiny{$0/\frac{1}{4}/\frac{1}{2}/\frac{3}{4}/1$} & Spawn distance from center (r) \\
Arena type & center     & Environment	\\
Eval. time  & 10    & Test duration in minutes \\ 
\bottomrule 
\toprule
\multicolumn{3}{l}{\textbf{Scalability}: \textit{Different swarm size}} \\ \midrule
Swarm size 			 & 10/20/50     & Number of robots in a swarm 	\\
Ratio      & 2:2/adaptive & sub-group division \\
Arena type 			 & center    & Environment 	\\
Eval. time  & 10    & Test duration in minutes \\ 
\bottomrule 
\toprule
\multicolumn{3}{l}{\textbf{Robustness}: \textit{Different environments}} \\ \midrule
Swarm size 			 & 20     & Number of robots in a swarm 	\\
Ratio      &2:2/adaptive & sub-group division \\
Arena type & bi\nobreakdash-mod./lin./ban.     & Environment 	\\
Eval. time  & 10    & Test duration in minutes \\ 
\bottomrule 
\end{tabular}
\vspace{-1.5em}
\label{tab:parameters_val}
\end{table}

\section{Results}\label{sec:res}
We provide full access to all the experimental data here after acceptance: \href{ANONYMOUS}{<ANONYMOUS placeholder>} %\href{https://doi.org/10.34894/0VSN8Z}{https://doi.org/10.34894/0VSN8Z}

\subsection{Evolutionary experiment}
\begin{figure}[h]
\vspace{-0.5em}
  \includegraphics[width=0.95\linewidth]{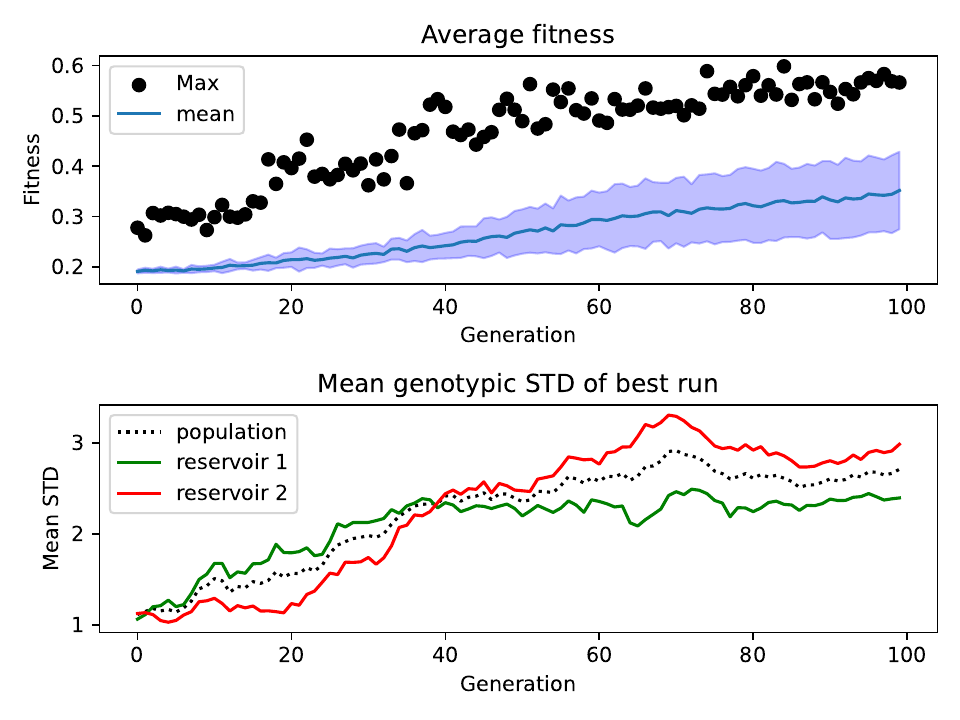}\vspace{-1.5em}
  \caption{\small Top: The mean$\pm STD$ (line) and average max (dot) fitness over 100 generations (averaged over 10 runs). Bottom: Mean genotype standard deviation (STD) within the population during the best run (per generation). The overall population mean as a black dotted line and each separate reservoir in solid red and green.}
  \label{fig:fitness}
  \vspace{-1.5em}
\end{figure}

We measure the efficacy by the mean and maximum fitness averaged over the 10 independent evolutionary runs for each generation. 
The results of our evolutionary experiment are presented in \autoref{fig:fitness}. We see a steady growth in mean fitness which indicates that the CMA-ES the collective gradient sensing task is learnable with our heterogeneous swarm. The maximum overall fitness is indicated by the black dots. We see a rapid increase around generation 20 which plateaus more or less around generation 50. 

At the bottom of \autoref{fig:fitness} we show the genotypical variation of our population in the best evolutionary run (calculated as mean Standard Deviation, STD). The rapid increase after generation 20 shows as an increase in variation, while the plateauing of fitness after generation 50 coincides with the steepest decline in genotype variation. It is interesting to see that both reservoirs seem to interchangeably adapt their genotype variation. Where the first reservoir (green) increases first but stabilizes relatively soon, while the second reservoir (red) remains relatively fixed at first and increases when the first reservoir starts to flatten out. This may indicate a concurrent adaptation of reservoirs due to a learned task distribution.

\subsection{Validation experiments}
% We retest our overall best controller. First, we are interested in emergent flocking behavior, which we investigate using our performance and order metrics (see \autoref{eq:r_i_sum} and \ref{eq:align}). 

\begin{figure*}[ht!]
    \centering
    % \begin{minipage}[c]{0.99\textwidth}
    % \raggedright\textbf{A} %## ROW 10x10
    % \end{minipage}
    \vspace{-1em}
    \begin{minipage}[c]{\textwidth}
        \begin{minipage}[t]{.16\textwidth}
            \centering
            \subfloat[1]{\includegraphics[width=\textwidth]{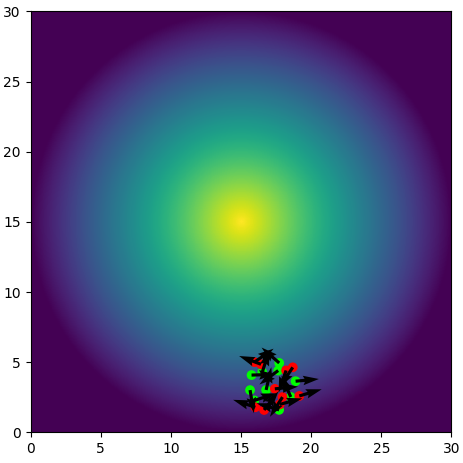}}
        \end{minipage}
        \hfill
        \begin{minipage}[t]{.16\textwidth}
            \centering
            \subfloat[2]{\includegraphics[width=\textwidth]{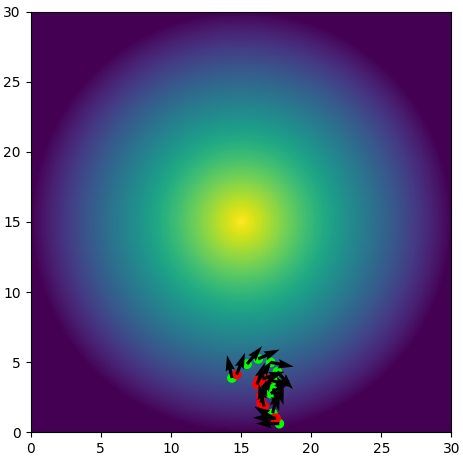}}
        \end{minipage}
        \hfill
        \begin{minipage}[t]{.16\textwidth}
            \centering
            \subfloat[3]{\includegraphics[width=\textwidth]{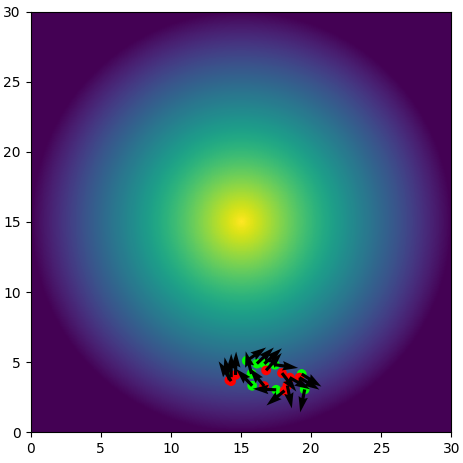}}
        \end{minipage}
        \hfill
        \begin{minipage}[t]{.16\textwidth}
            \centering
            \subfloat[4]{\includegraphics[width=\textwidth]{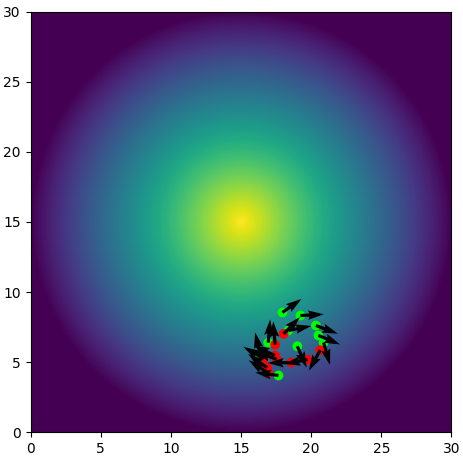}}
        \end{minipage}
        \hfill
        \begin{minipage}[t]{.16\textwidth}
            \centering
            \subfloat[5]{\includegraphics[width=\textwidth]{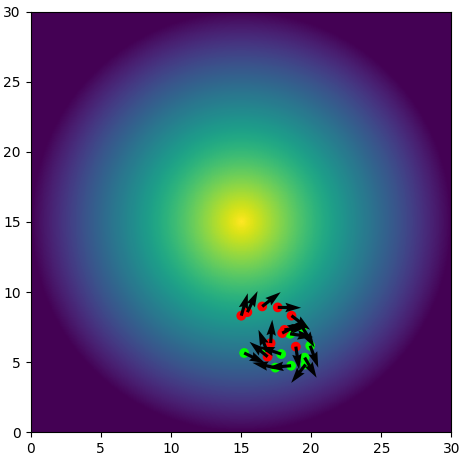}}
        \end{minipage}
        \hfill
        \begin{minipage}[t]{.16\textwidth}
            \centering
            \subfloat[6]{\includegraphics[width=\textwidth]{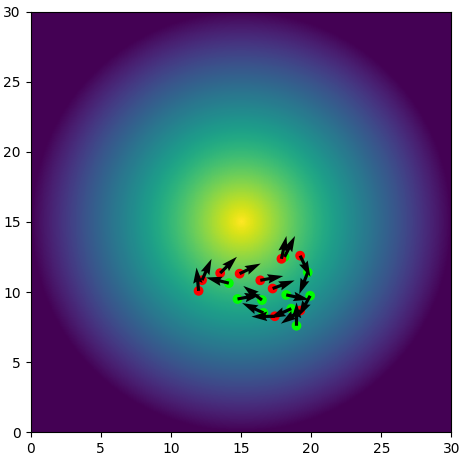}}
        \end{minipage}
\end{minipage}
\vspace{-1.5em}
    \caption{\small Snapshots of the interesting timeframes during the re-test experiment. The numbers correspond to the order of the vertical bars in the figure, corresponding to each line plot in \autoref{fig:retest}. (1) At first, the swarm spreads out to search for the gradient. (2) The green sub-group senses a gradient and 'aligns' the swarm to the left. (3) The gradient is lost, resulting in the swarm dispersing in different directions (i.e. decreasing alignment). (4) Red sub-group senses the gradient and directs the swarm towards the light source. (5) Red slowly pulls in more green swarm members. (6) the swarm starts to spread around the light source. }
    \label{fig:order}\vspace{-1.5em}
\end{figure*}

\begin{figure}[h]
% \vspace{-0.5em}
  \includegraphics[width=0.95\linewidth]{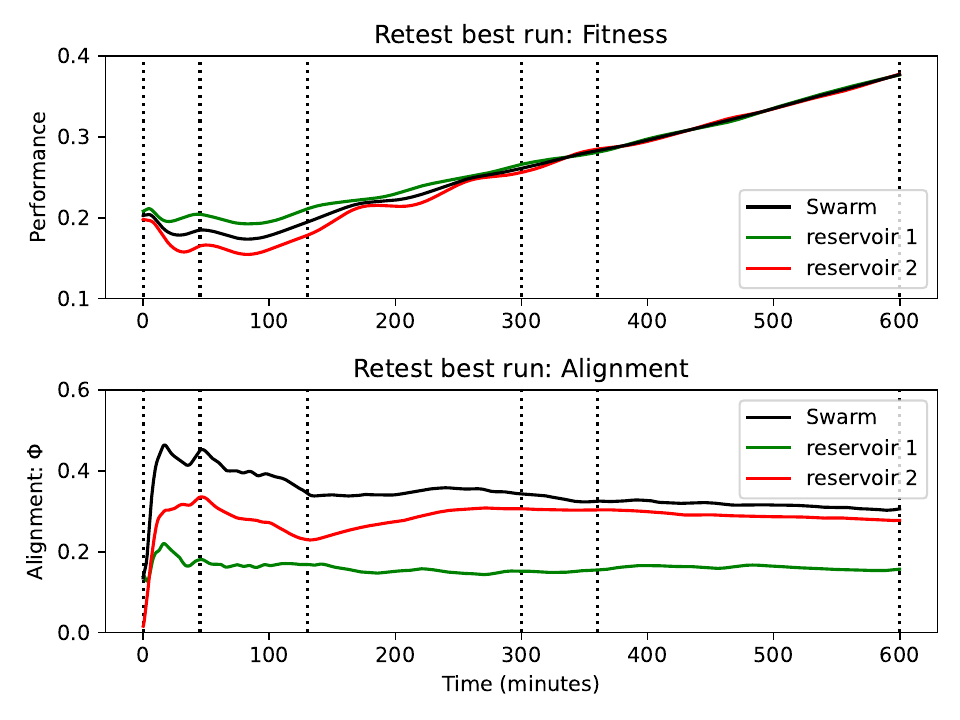}\vspace{-1.5em}
  \caption{\small Flocking behavior of the final best controller re-tested in the same environment. The three line plots on the left show the fitness (i.e. mean scalar light value) over time on the top, and the order (alignment) at the bottom. Vertical lines correspond with interesting time frames in \autoref{fig:order}.}
  \label{fig:retest}
\vspace{-1.5em}
  \end{figure}

\subsubsection{Collective behavior \& sub-group interactions}
\autoref{fig:order} and \ref{fig:retest} show the results of our best controller re-tested in the same environment. A video of this run is provided in the supplementary material (and is available here\footnote{\href{ANONYMOUS}{<ANONYMOUS placeholder>}}). During this trial we can measure fitness over time, with its final value around $0.38$. More interestingly, we measure the overall alignment of the swarm and the alignment for each sub-group. We provide snapshots of the swarm in the arena in \autoref{fig:order}, with the corresponding time frames represented as dotted vertical lines in \autoref{fig:retest} (the snapshot from left to right reflect the progression in time). 

In \autoref{fig:retest}, at the start, we see a low initial swarm alignment (black line) that quickly increases (corresponding to \autoref{fig:order}: 1-2). This rapid increase is mainly caused by sub-group 2 (in red) who tends to align more amongst themselves during the whole run. Sub-group 1 (in green) does not align that much, but finds the gradient faster at the start (see snapshot: 3), indicated by a higher fitness. Red follows shortly as a group given their alignment increase. In a later stage, red dominates the swarm's alignment and behavior, thereby concurrently pulling and pushing green towards the center (corresponding to \autoref{fig:order}: 4-5). This is also visible by the oscillating sub-group performance.

In \autoref{tab:ratio_robustness} we evaluated the same best swarm in different setups; we repeat the evaluation at different ratios of sub-groups and at different starting distances from the gradient center. Every configuration was repeated $60$ times.
We observe how starting closer to the target results in higher fitness.
We find that evenly mixed sub-groups at distances far away from the center ($r_{dist}\ge0.5$) perform statistically significantly higher than any of the extremes (i.e. fully green/red, $p\le0.05, df=118$).

When visually inspecting the behavior of the swarms at $r_{dist}=0$, we observe how both extreme ratios of sub-groups are capable of staying at the center of the gradient, but with different strategies (video in supplementary material and here\footnote{<ANONYMOUS placeholder>}). While the swarm made of only red robots seem to circle around the center with some distance, the swarm made of only green robots seem to fully occupy the space around the center, coinciding with a higher fitness. When evaluating the sub-group ratios at their extremes, we see that both independent sub-groups consistently outperform the best controllers of the first generation in our evolutionary experiments ($\sim 0.3$ at $r_{dist}=1.0$). However, a mixed ratio of subgroups performs better for all $r_{dist}\ge0$, indicating an advantage of sub-group interaction. Furthermore, we see a tendency of the second red sub-group to correlate with a higher performance at $r_{dist}\ge1.0$.

\newcommand*{\MinNumber}{0.05}
\newcommand*{\MidNumber}{0.7}
\newcommand*{\MaxNumber}{0.8}
\newcommand{\mytextcolor}[2]{\ifdim#1pt>\MidNumber pt\textcolor{gray!30}{#1}\else #1\fi}
\newcommand{\circletextcolor}[2]{\ifdim#1pt>\MidNumber pt\textcolor{gray!30}{\circlered{#2}{#1}}\else \circlered{#2}{#1}\fi}

\newcommand{\ApplyGradient}[2]{
    \FPeval{\result}{100*(#1-\MinNumber)/(\MaxNumber-\MinNumber)}
    \centering
    \edef\x{\noexpand\cellcolor{black!\result}}
    \x\mytextcolor{#1}
}
\newcommand{\ApplyGradientCircle}[2]{
    \FPeval{\result}{100*(#1-\MinNumber)/(\MaxNumber-\MinNumber)}
    \centering
    \edef\x{\noexpand\cellcolor{black!\result}}
    \x\circletextcolor{#1}{#2}
}
\newcommand{\ApplyGradientCircleA}[2]{\ApplyGradientCircle{#1}{a}}
\newcommand{\ApplyGradientCircleB}[2]{\ApplyGradientCircle{#1}{b}}
\newcommand{\ApplyGradientCircleC}[2]{\ApplyGradientCircle{#1}{c}}
\newcommand{\ApplyGradientCircleD}[2]{\ApplyGradientCircle{#1}{d}}
\newcommand{\ApplyGradientCircleE}[2]{\ApplyGradientCircle{#1}{e}}
\newcommand{\ApplyGradientCircleF}[2]{\ApplyGradientCircle{#1}{f}}
\newcolumntype{A}{>{\collectcell\ApplyGradientCircleA}p{0.5cm}<{\endcollectcell}}
\newcolumntype{B}{>{\collectcell\ApplyGradientCircleB}p{0.5cm}<{\endcollectcell}}
\newcolumntype{C}{>{\collectcell\ApplyGradientCircleC}p{0.5cm}<{\endcollectcell}}
\newcolumntype{D}{>{\collectcell\ApplyGradientCircleD}p{0.5cm}<{\endcollectcell}}
\newcolumntype{E}{>{\collectcell\ApplyGradientCircleE}p{0.5cm}<{\endcollectcell}}
\newcolumntype{F}{>{\collectcell\ApplyGradientCircleF}p{0.5cm}<{\endcollectcell}}
\newcolumntype{R}{>{\collectcell\ApplyGradient}p{0.5cm}<{\endcollectcell}}
\newcolumntype{Q}{>{\collectcell\ApplyGradient}p{0.4cm}<{\endcollectcell}}
\begin{table}[ht]
  \centering
  \renewcommand{\arraystretch}{1.6}
  \small
  % \begin{tabular}{ *{8}{Q} }
  %   0.1 & 0.2 &
  %   0.3 & 0.4 & 0.5 &
  %   0.6 & 0.7 & 0.8 \\
  %   \multicolumn{1}{c}{\vspace{-.7em}} \\
  % \end{tabular}
  \begin{tabular}{
    c  *{5}{R}
  }
  \toprule
      \multicolumn{1}{c}{ratio~$\rightarrow$}& 
      \multicolumn{1}{c}{\cellcolor{green!30}$4:0$} &
      \multicolumn{1}{c}{$3:1$} & 
      \multicolumn{1}{c}{$2:2$} & 
      \multicolumn{1}{c}{$1:3$} &
      \multicolumn{1}{c}{\cellcolor{red!30}$0:4$} \\
      % & $p<0.05$ \\
    \midrule
    % $r_{dist}=0.00$ & 0.70 & 0.74 & 0.77 & 0.78 & 0.79 \\
    % $r_{dist}=0.25$ & 0.68 & 0.73 & 0.75 & 0.76 & 0.76 \\
    % $r_{dist}=0.50$ & 0.64 & 0.67 & 0.70 & 0.71 & 0.70 \\
    % $r_{dist}=0.75$ & 0.50 & 0.56 & 0.60 & 0.60 & 0.57 \\
    % $r_{dist}=1.00$ & 0.36 & 0.44 & 0.45 & 0.42 & 0.34 \\

    % switched values left to right
    % $r_{dist}=0.00$ & \multicolumn{1}{A}{0.79} & 0.78 & 0.77 & 0.74 & 0.70 \\
    % $r_{dist}=0.25$ & 0.76 & \multicolumn{1}{B}{0.76} & 0.75 & 0.73 & 0.68 \\
    % $r_{dist}=0.50$ & 0.70 & \multicolumn{1}{C}{0.71} & 0.70 & 0.67 & 0.64 \\
    % $r_{dist}=0.75$ & 0.57 & 0.60 & \multicolumn{1}{D}{0.60} & 0.56 & 0.50 \\
    % $r_{dist}=1.00$ & 0.34 & 0.42 & \multicolumn{1}{E}{0.45} & 0.44 & 0.36 \\
    % $r_{dist}=1.25$ & 0.06 & 0.08 & \multicolumn{1}{F}{0.15} & 0.09 & 0.10 \\

    % 60 runs
    
    $r_{dist}=0.00$ & \multicolumn{1}{A}{0.79} & 0.78 & 0.77 & 0.73 & 0.69 \\% & \cellcolor{green!30}*** \\ % p=4.963e-23
    $r_{dist}=0.25$ & \multicolumn{1}{B}{0.76} & 0.76 & 0.75 & 0.72 & 0.68 \\% & \cellcolor{green!30}*** \\ % p=2.368e-23
    $r_{dist}=0.50$ & 0.70 & \multicolumn{1}{C}{0.71} & 0.70 & 0.66 & 0.64 \\% & \cellcolor{green!30}*** \\ % p=1.956e-08
    $r_{dist}=0.75$ & 0.56 & 0.60 & \multicolumn{1}{D}{0.60} & 0.57 & 0.52 \\% & \cellcolor{green!30}*   \\ % p=0.0108
    $r_{dist}=1.00$ & 0.33 & 0.41 & \multicolumn{1}{E}{0.43} & 0.43 & 0.38 \\% & \cellcolor{red!30}*   \\ % p=0.0116
    $r_{dist}=1.25$ & 0.06 & 0.08 & \multicolumn{1}{F}{0.13} & 0.09 & 0.12 \\% & \cellcolor{red!30}**  \\ % p=0.0015
    \bottomrule            
  \end{tabular}
  \begin{tikzpicture}[remember picture,overlay]
    % \draw[red,thick] (a) circle[x radius=4mm,y radius=2mm];
    \draw[red,thick] (a) ++(-4.1mm,-2.4mm) rectangle ++(8.2mm,5.3mm);
    \draw[red,thick] (b) ++(-4.1mm,-2.4mm) rectangle ++(8.2mm,5.3mm);
    \draw[red!70, dashed] (b) ++(4.4mm,-2.4mm) rectangle ++(8.2mm,5.3mm);
    \draw[red,thick] (c) ++(-4.1mm,-2.4mm) rectangle ++(8.2mm,5.3mm);
    \draw[red!70, dashed] (c) ++(4.4mm,-2.4mm) rectangle ++(8.2mm,5.3mm);
    \draw[red,thick] (d) ++(-4.1mm,-2.4mm) rectangle ++(8.2mm,5.3mm);
    \draw[red!70, dashed] (d) ++(-12.6mm,-2.4mm) rectangle ++(8.2mm,5.3mm);
    \draw[red,thick] (e) ++(-4.1mm,-2.4mm) rectangle ++(8.2mm,5.3mm);
    \draw[red,thick] (f) ++(-4.1mm,-2.4mm) rectangle ++(8.2mm,5.3mm);
    \draw[red!70, dashed] (f) ++(13mm,-2.4mm) rectangle ++(8.2mm,5.3mm);
  \end{tikzpicture}
  \hspace{1mm}
  % \begin{tabular}{ Q }
  %   % 0.1 \\ 0.2 \\
  %   0.05 \\ 0.2 \\ 0.4 \\
  %   0.6 \\ 0.8 \\ 
  % \end{tabular}
  % \vspace{-0.5em}
  \caption{\small Average fitness values ($N=60$) of retesting the best swarm with different sub-group ratios (green:red) from different starting distances ($r_{dist}$ = distance to the optimum). Sub-group ratios vary from solely sub-group 1 (green) to solely sub-group 2 (red). %Deviation from the mean varies from $0.01$ to $0.14$, depending on the configuration. 
  Solid red boxes indicate best ratio at a given $r_{dist}$, while the dashed boxed indicate no statistically significant differences with respect to the maximum.
    \label{tab:ratio_robustness}
  }
  \vspace{-2em}
\end{table}

\subsubsection{Online regulatory mechanism}
Based on the results of \autoref{tab:ratio_robustness} we assign the light intensity threshold values to the best performing sup-group ratio heuristically. I.e., we design a probabilistic state machine such that members in the swarm adapt their behavior automatically to reflect the optimal ratio on a swarm level using only local information. Thresholds are based on the light intensities at $r_{dist} = \{0.125, 0.375, 0.625, 0.875\}$. This results in the following function for the probability of expressing the first green sub-group behavior ($P_{green}$): 

\[   
P_{green}(light) = 
     \begin{cases}
       \text{1.0} &\quad\text{if $light$} >229\\
       \text{0.75} &\quad\text{if $light$} \in (76,229] \\
       \text{0.50} &\quad\text{if $light$} \le76\\
     \end{cases}
\]

Here, $light$ refers to the current light intensity measured by the robot ([0, 255]) and $P_{red} = 1 - P_{green}$.

\subsubsection{Scalability \& Robustness}
We test the best swarm controller with and without our online regulatory mechanism in 6 different environments (3 scalability experiments and 3 robustness). The results of these experiments are presented below (see \autoref{tab:res_val}). For \textit{Scalability}, the performance of the best controller seems to be positively correlated by the size of the swarm which indicates sensitivity to swarm size. This positive correlation is less apparent in the adaptive controller whose performance is statistically significantly higher for each swarm size comparison (10: $p\le0.05$, 20: $p\le0.01$, 50: $p\le0.001$). \textit{Robustness} results show the same tendency for the adaptive controller to outperform the controller without a regulatory mechanism, although these differences were not statistically significant. On aggregate, our online regulatory mechanism outperforms the best controller ($N=360$): Best $0.40\pm0.18$ \textit{vs.} Adaptive $0.45\pm0.22$. The aggregated means are statistically significantly different with Bonferroni correction ($p\le0.01/\alpha$ where $\alpha=6$, $df=718$).

\begin{table}[h]
    \centering
    \begin{tabular}{c ccc}
       \toprule
       \textbf{Scalability} & \multicolumn{3}{c}{\textit{Swarm size}}\\
       N=60  &  10 & 20 & 50\\
         \midrule
       Best  &  $0.35\pm0.10$ & $0.39 \pm 0.12$ & $0.47 \pm 0.03$\\
       
       Adaptive  &  $\hphantom{^*}\bm{0.40\pm0.12}^*$ & $\hphantom{^*}\bm{0.43\pm0.077}^*$ & $\hphantom{^*}\bm{0.49\pm0.03}^*$ \\
       \bottomrule
       \toprule
       \textbf{Robustness} & \multicolumn{3}{c}{\textit{Arena type}}\\
       N=60 &  \small{Bi-modal} & \small{Linear} &  \small{Banana} \\
       \midrule
       Best  &  $0.43 \pm 0.17$ & $0.51 \pm 0.19$ & $0.25 \pm 0.29$\\
       
       Adaptive  &  $\bm{0.47 \pm 0.19}$ & $\bm{0.59 \pm 0.25}$ & $\bm{0.31 \pm 0.35}$ \\
       \bottomrule
       
    \end{tabular}
    \caption{\small Validation experiments}
    \label{tab:res_val}
\vspace{-2em}
\end{table}

\section{Discussion}\label{sec:disc}
We successfully evolved self-organized sub-group specialization in a swarm of robots, a difficult task due to the complexity of swarm dynamics. This becomes even more evident when factoring in the added complexity of optimizing between-group interactions in conjunction with the specific specialization itself. The strength of our framework lies in its simple approach to learning these behaviors, which lends itself to broad applicability.
We demonstrated our methods' effectiveness in the context of an emergent perception task for gradient sensing, which is similar to most source localization tasks common in the swarm robotics literature.
Transferring our method to more complex tasks is also trivial, as 1) the use of task-agnostic nature of RNNs can express a wide array of behaviors, thus requiring no controller design adaptation; and 2) the simplicity of our group-level fitness function can be easily adapted, as its design requires minimal insight into the optimal solution (i.e. no presupposition on certain behaviors or specific task divisions are required).
Our general evolutionary approach to learn `sub-group specialization' is thus extremely useful for any type of task. 
% Furthermore, the general evolutionary approach to learn `sub-group specialization' can be extremely useful for any type of task especially considering the complex design of between-group interactions besides optimising for the specific specialization within the group itself. 
% Overall performance is comparable with previous work by <ANONYMOUS PLACEHOLDER>, although our current work shows more robust collective behavior in terms of alignment. This might be caused by the possibility for evolving specialization.

The results of our validation experiments show the emergence of specialization in our sub-groups. Sub-group 2 (red) seems to follow the gradient better at low intensities than sub-group 1 (see start of \autoref{fig:retest}) and tends to move more in coordination (i.e. higher alignment).
% This is also found in \autoref{tab:ratio_robustness}, where at $r_{dist}=1.0$ the ratio $4:0$ (i.e. sub-group 1) performs better than $0:4$. 
In contrast, sub-group 1 performs better when initialized near the center, and shows consistently lower alignment. This indicates that sub-group 1 shows less sensitivity to the swarm's overall behavior and tend to be more greedy as a sub-group. In contrast, sub-group 2 shows more exploratory behavior which provides more coordinated movement of the swarm when the gradient is found. 
% This more exploitative behavior of sub-group 2 also shows when we initialize our swarm at the center of the arena (\autoref{tab:ratio_robustness} $r_{dist}=0.0$).

Exploration and exploitation are fundamental principles in optimization, in general. You could interpret our task as such, where we learn our swarm to act as an optimizer that maximizes its local light value. The behavioral findings mentioned above show that the solution converged on a similar exploitation-exploration task division within their sub-group specializations. 
This task division is interesting as we did not encourage such collective behavior or specialization in our fitness. Arriving at these fundamental optimization principles in the context of sub-group swarm interactions without prior knowledge shows the power of automated design for finding collective behaviors suitable for a user-defined task.

From \autoref{tab:ratio_robustness} we can see that only employing one of the two sub-group at lower light intensities leads to lower performances than using a mixed ratio. This shows that the sub-group performance is enhanced by interactions with the other specialization. 
Collaboration becomes unnecessary when robots are placed near the center of the gradient ($r_{dist}\le 0.25$). 
The online regulatory mechanism furthermore shows the successful division of specialized tasks within our swarm, as it significantly improves performance of the best controller. The idea of biasing evolved specializations of the swarm adaptively toward a certain phenotype can be found in nature. Phenotypic plasticity emphasizes the expression of specific parts of the genome (i.e. our reservoirs) to obtain higher task competency on a swarm level. In our experiments, we successfully showed that this straightforward implementation of phenotypic plasticity results in higher scalability and significant overall performance.

\section{Conclusion}\label{sec:conc}
% specialized behavior in robot swarms can improve the overall swarm behavior in terms of efficacy and robustness. While this is a promising option in theory, designing controllers for a homogeneous swarm is already difficult, and designing controllers for sub-groups in the swarm is even harder due to the additional complexity of sub-group interaction.
Incorporating specialized behaviors within robot swarms holds the potential to significantly enhance overall swarm efficacy and robustness. However, this endeavor presents formidable challenges. Designing controllers for homogeneous swarms is inherently complex, and extending this to sub-groups within the swarm compounds the difficulty due to the added intricacy of sub-group interactions. In this paper, we show a viable approach to solve this challenge in a (sub)task-agnostic way, by co-evolving controllers heterogeneous swarm controllers while only specifying group-level task performance. We demonstrate that our evolved controllers show clear specialized sub-group behavior with sub-group interactions that improve the collective behavior. Furthermore, we show that learned sub-group behaviors can be used dynamically as an online regulatory mechanism to enhance performance and scalability. 

In the future, we propose to extend our work to encompass a broader spectrum of tasks, which could reveal other emergent specializations (communication, line-following, and mapping). Additionally, we foresee further improvements in the creation of more sophisticated controller designs where we can evolve the number of task-specializations and possibly the online adaptation rules to regulate phenotypic plasticity. Finally, we would like to test our controllers in a real world application for which we require the development of sensors to match our work. Such a milestone could also be a first step for realizing hardware-only swarm evolution experiments.
%This is different from conventional methods that take a more heuristic approach where they 1) analyse collective motion in animals; 2) designs controllers/rules to describe that motion; 3) optimize their controller rules to mimic the group behavior. Notably, this designs emergent behavior that is optimal for the task itself (whereas animal collective motion is optimized for everything that influences real wold survival). Secondly, human design of controllers/rules can be slow and limited (besides it being a tedious endeavor human design is often biased). Thirdly, our strategy makes more sense from a natural perspective, as an animal swarm optimizes for a specific task which consequently leads to group behavior as an optimal emergent property (i.e. optimization on task performance) and does not optimize for a specific group behavior itself (i.e. optimization on group behavior).

% \begin{acks}
% This work is supported by Technology Innovation Institute (TII).
% \end{acks}

%%%%%%%%%%%%%%%%%%%%%%%%%%%%%%%%%%%%%%%%%%%%%%%%%%%%%%%%%%%%%%%%%%%%%%%%

%%% The next two lines define, first, the bibliography style to be 
%%% applied, and, second, the bibliography file to be used.

\bibliographystyle{ACM-Reference-Format}
\bibliography{acmart}

\end{document}